\documentclass[letterpaper]{article} 
\usepackage{aaai24}  
\usepackage{times}  
\usepackage{helvet}  
\usepackage{courier}  
\usepackage[hyphens]{url}  
\usepackage{graphicx} 
\usepackage{natbib}  
\usepackage{caption} 
\usepackage{algorithm}
\usepackage{algorithmic}

\usepackage{amsmath}
\usepackage{multirow}
\usepackage{subcaption}

\usepackage{newfloat}
\usepackage{listings}
\DeclareCaptionStyle{ruled}{labelfont=normalfont,labelsep=colon,strut=off} 
\floatstyle{ruled}
\newfloat{listing}{tb}{lst}{}
\floatname{listing}{Listing}
\newcommand{\country}[1]{\textit{#1}}

\title{A Temporal Graph Network Framework for Dynamic Recommendation}
\author{
    Yejin Kim\textsuperscript{1},
    Youngbin Lee\textsuperscript{1},
    Vincent Yuan\textsuperscript{2},
    Annika Lee\textsuperscript{2},
    Yongjae Lee\textsuperscript{1}\thanks{Corresponding author.}
}
\affiliations {

    \textsuperscript{\rm 1}Ulsan National Institute of Science and Technology, \country{South Korea}\\
    \textsuperscript{\rm 2}Carnegie Mellon University, \country{United States}\\

    \{kimyejin99, young, yongjaelee\}@unist.ac.kr, \{jyuan2, juhyeonl\}@andrew.cmu.edu

}

\usepackage{bibentry}

\begin{document}

\maketitle

\begin{abstract}
Recommender systems, crucial for user engagement on platforms like e-commerce and streaming services, often lag behind users' evolving preferences due to static data reliance. After Temporal Graph Networks (TGNs) were proposed, various studies have shown that TGN can significantly improve situations where the features of nodes and edges dynamically change over time. However, despite its promising capabilities, it has not been directly applied in recommender systems to date. Our study bridges this gap by directly implementing Temporal Graph Networks (TGN) in recommender systems, a first in this field. Using real-world datasets and a range of graph and history embedding methods, we show TGN's adaptability, confirming its effectiveness in dynamic recommendation scenarios.
\end{abstract}

\section{Introduction}
In the rapidly evolving digital world, recommender systems have become a cornerstone of user experience, profoundly influencing choices in e-commerce, content streaming, and social networking platforms \cite{9118884}. 
As interaction-based recommender systems started to emerge, models such as NGCF \cite{wang2019neural} and LightGCN \cite{he2020lightgcn} that include neighborhood aggregation features but also linearly propagate the embeddings on the user-item interaction graph have been widely used. 
However, these existing models often struggle to adequately capture temporal variations and the continuous evolution of user-item interactions and features due to their static nature. These static models are unable to accommodate the temporal changes in user preferences, leading to outdated and less relevant recommendations \cite{Gao_2021}. \\

Addressing the need to capture evolving user preferences over time, sequential recommendation models \cite{tan2016improved,sun2019bert4rec,kang2018self} have emerged. These models utilize users' purchase sequences, employing attention mechanisms in recent advancements for next-item prediction. Despite their efficiency, these models face challenges in adapting to the dynamic nature of user-item interactions, prompting the exploration of dynamic recommendation models. Moreover, while models \cite{bai2020temporal, xiang2010temporal} use the term 'temporal' in their names, they incorporate temporal aspects only in a rudimentary fashion. These approaches typically treat time as a linear or categorical feature, overlooking the nuanced and evolving relationships between users and items. \\


Amidst these developments, Twitter proposed a framework called Temporal Graph Networks (TGN) \cite{rossi2020temporal}, suitable for dynamic situations.
TGN shows promising effects in different aspects. Its history module provides the exceptional capability of TGN to keep long-term history dependencies for each node in the graph. This means that nodes can be kept updated based on historical data without continuous training, which is essential for dynamic change over the network. In addition to this, its graph embedding allows to compute the up-to-date embedding for a node by aggregating one’s neighbor nodes’ memories even if a node has been inactive for a while.
TGN can be used for various tasks such as edge prediction and node classification. The authors have mentioned that TGN has potential future research directions in social sciences, recommender systems, and biological interaction networks. \\

Zhao et al. \cite{zhao2023time} highlight the potential of TGN. They proposed a time-interval aware recommendation model using bi-directional continuous time dynamic graphs, employing TGN as a baseline. Their findings indicate that TGN variants, leveraging historical history messages, outperform other baselines, underscoring TGN's suitability for dynamic recommendations. However, the paper's scope is limited, as it only utilizes JODIE \cite{kumar2018learning} and DyRep \cite{trivedi2019dyrep} among TGN's variations, without detailing their specific applications or explaining how these advantages impact the recommendation system. Consequently, to the best of our knowledge, a direct application of the TGN framework within recommender systems emerges as a novel and unexplored approach.\\

\begin{figure*}[htp]  
    \centering
    \includegraphics[width=\textwidth, height=0.8\textheight, keepaspectratio]{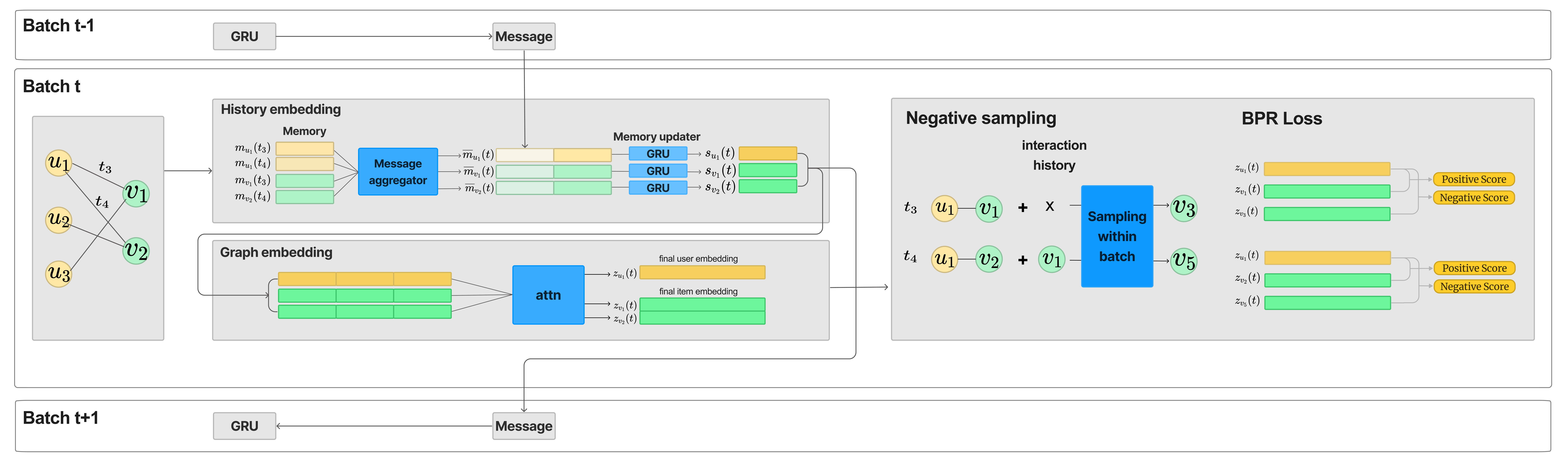}
    \caption{Overview of TGN framework for dynamic recommendation}
    \label{fig:data}
\end{figure*}

\indent In light of this research gap, our study undertook the direct implementation of the TGN framework within a recommender system setting, aiming to fully harness its potential in capturing dynamic user-item interactions over time. In particular, we suggested different variations across history embedding and graph embedding methods. 
This approach marks a significant step forward in the development of more sophisticated and temporally aware recommender systems. 
Our key contributions are as follows:
\begin{itemize}
    \item We proposed a TGN framework for dynamic recommendation tailored for use in recommender systems. Its primary feature is the ability to learn from user-item interactions over time, offering a more realistic and updated understanding of user preferences. 
    \item We tested our TGN framework using real-world datasets. The positive results from these tests confirm the framework's effectiveness in practical scenarios.
    \item To our knowledge, this is a new direction in this field of study. The significance of this work lies in its exploration of how TGN can be applied directly in recommender systems, a concept not widely explored before. 
\end{itemize}

\section{Preliminaries}
\subsection{Problem Definition}
We introduce the definition of the task associated with dynamic recommendations. For the problem in focus, let’s denote the user set as $U = \{u_1, u_2, ... u_{|U|}\}$, item set as $V = \{v_1, v_2,...,v_{|V|}\}$, and time set as $T = \{t_1, t_2, ... t_{|T|}\}$. Then, the interaction can be represented by $y_{u,v}^t$. If user $u$ interacts the item $v$ at time $t$, then $y_{u,v}^t=1$; otherwise $y_{u,v}^t=0$. 

Our primary goal is to predict the value of $y^{t}_{u,v}$. Ultimately, for each user and time instance, the system aims to select and recommend the top-k items, leading to a personalized and time-sensitive set of recommendations for each user in the system.

\subsection{Continuous Time Dynamic Graph}
The graph of interest in our problem is not static but dynamic, changing its structure over continuous time. This necessitates a representation suitable for temporal variations. Therefore, we define a continuous-time bipartite graph as $\mathcal{G}(T)=(\mathcal{V}, \mathcal{E}_T)$ where $\mathcal{V}$ denotes $U,V \in \mathcal{V}$ and $\mathcal{E}_T$ denotes the set edges at time t. Each edge in $\mathcal{E}_T$ is characterized by a tuple $e=(u,v,t, \mathbf{e}_{uv})$, consisting of a start node (user), an end node (item), a timestamp $t$, and an edge feature $\mathbf{e}_{uv}$. 
This stands apart from sequential recommendations by integrating edge features that dynamically evolve over time, enhancing its ability to capture dynamic user behavior and item features.

\section{Methodology}
We propose the TGN Framework, inspired by \cite{rossi2020temporal}, for a recommender system that is designed to handle temporal dynamics and interactions between users and items. Our goal is to effectively learn the evolving characteristics of users and items over time using graph-based methods. To achieve this, we introduce an effective temporal graph embedding method, known as TGN, along with a recommendation approach that utilizes this method.



\subsection{Temporal Graph Networks}


\subsubsection{Memory embedding}

The approach begins with generating memory embeddings for each node to capture their dynamics. We extract information associated with a node $i$ at time $t$ as $m_{i}(t)  = s_i\left(t^{-}\right) \| s_j\left(t^{-}\right) \| \Delta t \| \mathbf{e}_{i j}$, where $\mathbf{s}_i\left(t^{-}\right)$ and $\mathbf{s}_j\left(t^{-}\right)$ represent the memory embedding at the previous time step for the source and destination nodes, respectively. $\Delta t$ is the time interval, and $\mathbf{e}_{ij}$ is the edge feature. In instances where nodes appear multiple times within the same batch, we incorporate information from the last time step to update the memory.


After extracting information, we use recurrent neural network to update memory embeddings. Using GRU \cite{chung2014empirical}, memory of node $i$ can be updated as follows: 


\begin{equation}
    s_i(t)=GRU(\overline{m}_i(t), s_i(t^-))
\end{equation}

\subsubsection{Graph embedding}

In this module, node embeddings are created for each node at time step $t$, resulting in node embeddings that encapsulate temporal information. If graph attention is utilized, a node embedding can be represented as:

\begin{equation}
    \mathbf{z}_i(t)=\sum_{j \in n_i^k(t)} attn\left(s_i(t), s_j(t), \mathbf{e}_{i j}\right)
\end{equation}

where $attn$ refers to the graph attention mechanism described in \cite{rossi2020temporal}, and $s$ represents memory. The neighborhood set of node $i$, denoted as $n_i^k(t)$, refers to the k-hop temporal neighborhood connected up to time $t$. Other methods such as temporal graph sum \cite{rossi2020temporal} and graph convolutional network (GCN) can be utilized for graph embedding learning.

\subsection{Model Optimization}

Our main contribution is proposing a suitable learning approach for recommendation within the TGN framework. In other words, we introduce the methods of negative sampling and loss formulation that enable recommend top-k items for each user.

\subsubsection{Negative sampling}
We have a unique setting that requires negative sampling, taking into account time. Instead of using a static positive item set, we utilize a positive item set consisting of items that the user has purchased up to time $t$ within a batch. In other words, given a user $u$ and a specific time $t$, $p_{u,t}$ represents user's current positive item set. Then, from all item sets within the batch, we randomly sample $n$ candidate items that do not belong to $p_{u,t}$. That is, we sample items from the set $V_B - p_{u,t}$ where $V_B$ is the set of items in a batch where $t$ belongs to.


\subsubsection{BPR loss}

At the time when the interaction takes place, an item interaction by a user $u$ treated as positive. The set of negative items is sampled for the same time point.  Then, Bayesian Personalized Ranking (BPR) loss is applied to scores for positive pairs and negative pairs.

\begin{equation}
    \mathcal{L}_{B P R}=\sum_{(u, p, n, t) \in D}-\log \sigma\left(z_u(t)^T z_p(t)-z_u(t)^T z_n(t)\right)
\end{equation}

In this equation, \(D\) denotes the edge set, which is derived from \(\mathcal{E}_T\) with additional negative sampling. Here, \(u\) represents user, \(p\) is positive item, \(n\) is negative item, and \(t\) represents time.

\begin{table*}[ht!]
\centering
\renewcommand{\arraystretch}{1.3} 
\resizebox{\textwidth}{!}{%
\begin{tabular}{|c|c|c|c|c|c|c|c|c|c|c|c|c|}
\hline
Dataset & Metric & ItemKNN & BPR & NGCF & LightGCN & GRU4Rec & STAMP & SASRec & Jodie & DyRep & TGN & \textit{improv.}\\ \hline
\multirow{3}{*}{MovieLens} & Recall@5 & 0.0494 & 0.0471 & 0.0444 & 0.0472 & 0.0455 & 0.0506 & \textbf{0.0584} & - & 0.0483& \underline{0.0577}& -1.19\% \\  \cline{2-13} 
 & Recall@10 & 0.0825 & 0.0727 & 0.0643 & 0.0807 & 0.0812 & 0.0897 & \underline{0.1021} & - & 0.0966& \textbf{0.1126}& 10.28\% \\ \cline{2-13} 
 & Recall@20 & 0.1338 & 0.1127 & 0.1044 & 0.1197 & 0.1379 & 0.1417 & 0.1561 & - & \underline{0.1916}& \textbf{0.2211}& 15.40\% \\ \hline
\multirow{3}{*}{retailrocket} & Recall@5 & 0.0197 & 0.0422 & 0.0713 & 0.0366 & 0.0454 & 0.0785 & \underline{0.0997} & 0.0145 & 0.0165& \textbf{0.1030}& 3.31\%\\ \cline{2-13} 
 & Recall@10 & 0.0480& 0.0824 & 0.1167 & 0.0518 & 0.0679 & 0.1022 & \underline{0.1428}& 0.0265 & 0.0295& \textbf{0.2000}& 40.02\%\\ \cline{2-13} 
 & Recall@20 & 0.1118 & 0.1830& 0.2276 & 0.1011 & 0.1242 & 0.1371 & \underline{0.2238}& 0.0540 & 0.0535& \textbf{0.3610}& 61.23\%\\ \hline
\end{tabular}
}
\caption{Performance comparison with baselines. The best scores in each row are represented in bold, while the second-best scores are underlined. The improvement values (\textit{improv.}) indicate TGN's percentage changes relative to the best-performing baseline.}
\label{tab:comparison_methods}
\end{table*}

\section{Experiments}
\subsection{Experiments Settings}
\subsubsection{Datasets}

To assess the model's ability to effectively capture temporal dynamics, we employed datasets in which interactions are ordered by time. The MovieLens dataset (1,000,209 interactions, 6,040 users, and 3,952 items) and the RetailRocket transaction dataset (22,457 interactions, 11,719 users, and 12,025 items) were utilized for this purpose. 

\subsubsection{Baseline}
To assess the effectiveness of our stock recommendation model, we compared it with basic recommendation models, static graph-based recommendation models, sequential recommendation models, and dynamic graph learning models. For the basic recommendation models, we utilized Pop, which recommends the most frequently traded items, ItemKNN \cite{deshpande2004item} based on item similarity, and BPR \cite{rendle2012bpr}, a matrix factorization model. Within the static graph-based recommendation models, we employed NGCF \cite{wang2019neural}, which uses graph convolutional networks (GCN) to learn node embeddings of the user-item bipartite graph, and its lightweight version, LightGCN \cite{he2020lightgcn}. For sequential recommendation models, we utilized SASRec \cite{kang2018self} and STAMP \cite{liu2018stamp}, both based on Transformer architectures, and GRU4REC \cite{tan2016improved} were used.  In the dynamic methods category, Jodie \cite{kumar2018learning} and DyRep \cite{trivedi2019dyrep} were used. 

\subsubsection{Hyperparameter Setting}
Our model was experimented with the following settings: 10 epochs, a batch size of 1000, a learning rate of 0.0001, a node embedding dimension of 31, a memory embedding dimension of 31, and a time embedding dimension of 100. Additionally, following the experimental results in  \cite{rossi2020temporal}, we used GRU as the memory embedding module and Graph Attention as the graph embedding module. For baseline models, the number of epochs was consistently set to 50 with a batch size of 2000 for each experiment.

\subsubsection{Metrics and Evaluation}
We utilized a commonly used metric, Recall, to assess the effectiveness of a ranked list. In the process of evaluating, we take 100 negative samples per each positive samples. Following the setting of dynamic graph learning model \cite{kumar2018learning}, we performed a chronological split of the data, dividing interactions into training, validation, and test sets in an 8:1:1 ratio based on the chronological order of time.


\subsection{Performance Comparison}
In our comprehensive evaluation, we focused on contrasting the effectiveness of static, sequential, and temporal recommender models across two key datasets: MovieLens and RetailRocket. The evaluation metrics centered on Recall at varying thresholds (5, 10, and 20).



The highlight of our analysis was the performance of our model, TGN. In both datasets, TGN demonstrated remarkable superiority in almost all evaluated metrics. For the MovieLens dataset, TGN achieved an impressive Recall@20 score of 0.2211, outshining both the static and sequential models. In the RetailRocket dataset, TGN's performance was even more striking, with a Recall@20 score of 0.3610, leading all other models.

The results of our study are particularly noteworthy given the intricacies and challenges involved in integrating temporal aspects into recommendation models. TGN's adeptness in incorporating time-aware components has proven pivotal in accurately capturing the dynamic and evolving nature of user preferences. This effectiveness underscores the vital importance of considering temporal dynamics in recommendation systems, especially pertinent in scenarios where user preferences are not static, but fluid and subject to change over time. Additionally, the effectiveness of GNN module used for learning the final graph embedding is evident. It successfully reflected user-item interactions, reinforcing the model's comprehensive approach. TGN's outstanding performance, despite the demanding predictive requirements of time-aware modeling, distinctly highlights its robustness and effectiveness, making it highly relevant in modern recommendation scenarios.

One important point to note is that among temporal models, Jodie and DyRep perform poorly. This poor performance seems to be attributed to their method of generating graph embeddings based on time and ID, respectively. This suggests that approaches not utilizing GNN modules are unsuitable for recommender system datasets forming user-item bipartite graphs. Therefore, GNN modules are crucial for the effectiveness of TGN frameworks in recommender systems. We also attempted a comparison with JODIE using the MovieLens dataset; however, we do not report it due to out-of-memory errors on this large dataset.


\subsection{Choice of Modules}


\begin{table}[ht] 
\centering
\renewcommand{\arraystretch}{1.3} 
\begin{tabular}{|c|c|c|c|c|}
\hline
\multirow{2}{*}{Dataset} & \multirow{2}{*}{Module} & \multicolumn{3}{c|}{Recall@10} \\ \cline{3-5} 
 &  & attn & sum & GCN \\ \hline
\multirow{2}{*}{Movielen} & GRU & 0.1126 & 0.1075 & 0.1033 \\ \cline{2-5} 
 & RNN & 0.1076 & 0.1053 & 0.1039 \\ \hline
\multirow{2}{*}{retailrocket} & GRU & 0.1890 & 0.1630 & 0.1330 \\ \cline{2-5} 
 & RNN & 0.2000 & 0.1760 & 0.1440 \\ \hline
\end{tabular}
\caption{Recall@10 metrics for different modules in Movielen and retailrocket datasets.}
\label{tab:module_choice}
\end{table}

The advantage of utilizing the TGN framework lies in the ability to choose from various embedding modules. To examine the effects of each module, an ablation study comparing them was performed. Specifically, experiments were conducted using GRU and RNN as history embedding methods and temporal graph attention (attn), temporal graph sum (sum), and graph convolution network (GCN) as graph embedding methods. Summarized results in Table \ref{tab:module_choice} reveal that among the graph embedding methods, the attn method demonstrated the best performance across all datasets. This is attributed to its ability to obtain the most recent information from the graph and select crucial neighbors effectively. While the TGN paper confirmed this through basic graph learning tasks such as node classification and edge prediction, this paper further confirms that these TGN characteristics are applicable to recommendation datasets and tasks. Regarding history embedding methods, different results were observed depending on the dataset. This indicates that the influence of module selection is not as pronounced as with graph embedding methods. Therefore, it is reasonable to select an appropriate method based on the dataset.

\section{Conclusion and Discussion}

In this study, we introduced the TGN framework for the dynamic recommender system. It demonstrates a significant improvement over traditional static models by accurately adapting to the dynamic nature of user preferences. The application of TGN captures the temporal shifts in user behavior, showcasing both theoretical and practical enhancements in real-world dataset evaluations.

Looking to the future, our focus will shift towards enhancing the practicality of TGN for large-scale applications. This will involve integrating a temporal network hashing \cite{vaudaine2023temporal} method, designed to optimize learning efficiency and reduce memory requirements while preserving the temporal sensitivity of the data. This advancement is pivotal for the scalable application of TGN-based recommender systems, potentially leading to more responsive and efficient platforms across various online services.

\section{Acknowledgments}
This work was supported by Institute of Information \& communications Technology Planning \& Evaluation (IITP) grant funded by the Korea government(MSIT) (RS-2022-00143911, AI Excellence Global Innovative Leader Education Program)

\bibliography{aaai24}

\end{document}